\pdfoutput=1

\documentclass[11pt]{article}

\usepackage[utf8]{inputenc} 
\usepackage[T1]{fontenc}    
\usepackage{hyperref}       
\usepackage{url}            
\usepackage{booktabs}       
\usepackage{amsfonts}       
\usepackage{nicefrac}       
\usepackage{microtype}      

\usepackage{amssymb}
\usepackage{amsbsy}
\usepackage{amsmath}
\usepackage{amsthm}
\usepackage{latexsym}
\usepackage{subcaption}
\usepackage{wrapfig}

\usepackage{graphicx}
\usepackage{tabularx}
\usepackage{ulem}
\usepackage[font=small]{caption}
\usepackage{algorithm}
\usepackage{algorithmicx, algpseudocode}
\usepackage{mathtools}
\usepackage{arydshln}
\usepackage{titlesec}
\usepackage{multirow}
\usepackage[dvipsnames]{xcolor} 
\usepackage[]{acl}
\usepackage{graphicx}
\usepackage{times}
\usepackage{latexsym}
\usepackage{booktabs}
\usepackage{multirow}
\usepackage{hyperref}
\usepackage{url}
\usepackage{amssymb}
\usepackage{amsmath}
\usepackage{xspace}
\usepackage{booktabs}
\usepackage{caption}
\usepackage{subcaption}
\usepackage{inconsolata}
\usepackage{tabularx}
\usepackage{xcolor}
\definecolor{urlcolor}{HTML}{C30004}

\usepackage{todonotes}
\usepackage[dvipsnames]{xcolor} 
\title{\textsc{MolBind}: Multimodal Alignment of Language, Molecules, and Proteins}

\author{Teng Xiao$^1$, Chao Cui$^2$, Huaisheng Zhu$^1$ \and Vasant G Honavar$^1$ \\
  $^1$Penn State University $^2$Tsinghua University \\
  \texttt{\{tengxiao, hvz5312, vhonavar\}@psu.edu, chaocui01@gmail.com}}

\begin{document}
\maketitle

\begin{abstract}
Recent advancements in biology and chemistry have leveraged multi-modal learning, integrating molecules and their natural language descriptions to enhance drug discovery. However, current pre-training frameworks are limited to two modalities, and designing a unified network to process different modalities (e.g., natural language, 2D molecular graphs, 3D molecular conformations, and 3D proteins) remains challenging due to inherent gaps among them. In this work, we propose \textsc{MolBind}, a framework that trains encoders for multiple modalities through contrastive learning, mapping all modalities to a shared feature space for multi-modal semantic alignment. To facilitate effective pre-training of \textsc{MolBind} on multiple modalities, we also build and collect a high-quality dataset with four modalities, \texttt{MolBind-M4}, including graph-language, conformation-language, graph-conformation, and conformation-protein paired data. \textsc{MolBind} shows superior zero-shot learning performance across a wide range of tasks, demonstrating its strong capability of capturing the underlying semantics of multiple modalities. Our code and data can be found at \href{https://github.com/tengxiao1/MolBind}{\color{urlcolor}{MolBind}}.



\end{abstract}

\section{Introduction}
\label{sec:intro}

\begin{figure}[t]
\centering
\includegraphics[width=0.35\textwidth]{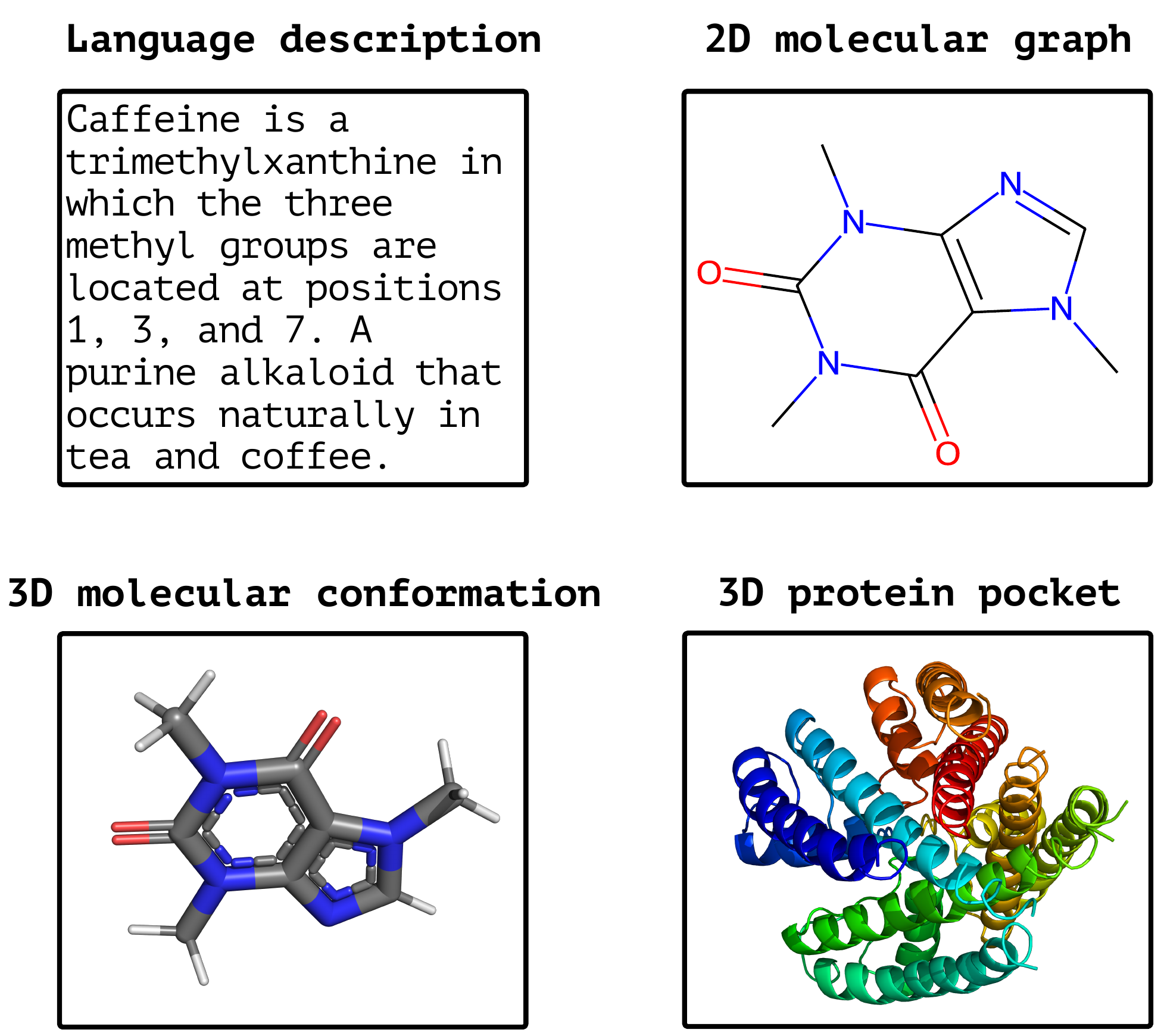}
\vskip -0.5em
     \caption{\small Illustration of modality data we try to align in this paper. The scientific language: text describes the name, property, and structure of molecules. 2D molecular graph: the atoms and bonds are nodes and edges in the graph. 3D molecular conformation: the atoms in the 3D Euclidean space. 3D protein pocket: a specific 3D region on the surface of a protein that can bind or interact with other molecules.}
\label{fig:data} 
\vskip -1.5em
\end{figure}

Multi-modal  pre-training, as exemplified by the contrastive language-image pre-training model CLIP~\cite{radford2021learning}, has made significant strides in computer vision and natural language processing. By aligning latent representations of images and textual descriptions, CLIP demonstrates remarkable zero-shot capabilities. This success  spurred the adoption of multi-modal pre-training in many domains, including audio~\cite{guzhov2022audioclip}, video~\cite{xu2021videoclip}, tabular data~\cite{hager2023best}, and point clouds~\cite{zhang2022pointclip}.

Remarkably, the properties of molecules can be effectively described by language, i.e., biomedical texts with rich expressions. Consequently, multi-modal pre-training has been adopted in specialized scientific fields such as biology, facilitating advancements in molecule-language alignment~\cite{li2024towards,su2022molecular,edwards2021text2mol}.  Molecule-language alignment
has brought  tremendous improvements for example for language-guided molecule generation~\cite{liu2023multi,zhu3MDiffusion}, molecule-language retrieval~\cite{edwards2021text2mol}, and molecule-based text generation~\cite{liu2023molca,li2024towards}.

Despite these successes, these approaches face two critical limitations: (i) The availability of multi-modal pairs for molecules is insufficient and significantly smaller compared to other domains. For example, CLIP is trained on a dataset of 400M image-language pairs collected from the internet, whereas the total number of available molecule-language pairs is only about 300K~\cite{liu2023multi,liu2023molca} which is orders of magnitude lower. Despite the achievements of multi-modal pre-training, its broader and more flexible application in biology domains remains limited by the high dependence on large-scale, high-quality paired data. (ii) Current multi-modal methods for molecules utilize only a single pair of modalities, limiting their extension to multiple N ($\geq 3$) modalities. Thus, the final representations are restricted to the pairs of modalities used during training. For instance, aligned molecule-language representations cannot be directly applied to conformation-protein tasks, and vice versa.

To address these challenges, we propose \textsc{MolBind}, a novel multi-modal pre-training framework that extends molecule-language pre-training to multiple modalities. Inspired by recent work~\cite{girdhar2023imagebind}, attempting to bind various image-paired datasets, we bind all modalities (\texttt{molecular language description}, \texttt{2D molecular graph}, \texttt{3D molecular conformation}, \texttt{3D protein pocket}) as shown in Figure~\ref{fig:data} to a unified embedding space, enabling effective semantic alignment and facilitating a comprehensive understanding of various modalities in the biology domain. 
\textsc{MolBind} does not require datasets where all modalities co-occur for each sample, which are very hard to collect. Instead, it learns a single alignment representation space by leveraging multiple types of modality pairs. To validate our \textsc{MolBind}, we build and collect a  high-quality dataset with \texttt{4} modalities, \texttt{MoBind-M4} including graph-language, conformation-language, graph-conformation, and conformation-protein paired data. By augmenting the alignment process with data from multiple modalities, \textsc{MolBind} enhances model robustness against data sparsity and yields discernible improvements in downstream tasks.

\textbf{Our key contributions are:}
\textbf{(i)} We propose a novel framework named \textsc{MolBind}, which effectively extracts shared semantic information from paired multi-modal data and gradually aligns various molecular modalities with the language modality into a joint space.
\textbf{(ii)} We collect and construct a multi-modal dataset for molecules, comprising data pairs of four modalities. To the best of our knowledge, \texttt{MoBind-M4} is the first unified dataset, containing multiple modalities of language, molecules, and proteins in the biological domain.
\textbf{(iii)} Extensive experiments on downstream tasks, such as zero-shot cross-modal retrieval and cross-modal classification, validate the effectiveness of \textsc{MolBind} in representation alignment between language and molecular modalities.

\section{Method}
\label{sec:method}
In this section, we present \textsc{MolBind}, a multi-modal pre-training method designed to align the latent semantic representations of multiple modalities in the biology domain with natural language.


\subsection{Language, Molecule and Protein Encoders}
In this section,  we briefly introduce multi-modal encoders we choose. However. 
\textsc{MolBind} is conceptually simple and diverse models can be used as language, molecule, and protein encoders.\\
\textbf{Language Encoder.}
The textual language modality delivers an extensive semantic analysis of the molecule’s structure, properties, and functionality. Such domain knowledge, in the form of natural language, is encoded using SciBERT~\cite{beltagy2019scibert}, which was pre-trained on textual data from the biological domain, following previous works~\cite{li2024towards,liu2023molca,liu2023multi}.\\
\textbf{2D Graph Encoder.}
To encode the modality of 2D molecular graph, we utilize graph neural networks~\cite{hu2020strategies,xiao2022decoupled,xiao2021learning}. Specifically, we use the pre-trained graph isomorphism network (GIN) encoder~\cite{hu2020strategies} which is pre-trained on 2 million molecules from the ZINC15~\cite{sterling2015zinc} using graph contrastive learning~\cite{you2020graph}.\\
\textbf{3D Conformation Encoder.}
We adopt Uni-Mol~\cite{zhou2022uni} as  3D conformation encoder (details in Appendix~\ref{sec:appencoder}). Specifically, Uni-Mol, a transformer-based model, has been pre-trained on 209 million molecular conformations and is capable of effectively capturing the input 3D geometry information of molecules.\\
\textbf{3D Protein Encoder.}
Similar to 3D conformation, we also architect the 3D protein encoder as a Uni-Mol  and initialize it from the checkpoint pre-trained on a 3M candidate protein pockets dataset. 

\noindent We employ separate encoders for language, 2D molecular graphs, 3D molecular conformations, and 3D protein pockets. On each encoder, we add a modality-specific linear projection head to obtain a fixed-size, $D$-dimensional embedding. This embedding is normalized and utilized in subsequent contrastive learning to bind individual modalities.

\subsection{Multi-modal Representation Alignment}
\textsc{MolBind} utilize various pairs of modalities $(\mathcal{X}, \mathcal{Y})$, where $\mathcal{X}$
represents one modality and $\mathcal{Y}$ another, to learn a single joint embedding. To ensure alignment across  modalities, we adopt multi-modal contrastive learning principles~\cite{radford2021learning}.  Given a sample $x_i$ and its corresponding observation in the other modality $y_i$, we encode them into normalized embeddings: $\mathbf{x}_i=f_{\mathcal{X}}\left({x}_i\right)$ and $\mathbf{y}_i=f_{\mathcal{Y}}\left(y_i\right)$ where $f_{\mathcal{X}}$ and $f_{\mathcal{Y}}$ are encoders corresponding to modalities $\mathcal{X}$  and $\mathcal{Y}$. The encoders are optimized to increase the similarity of paired data, align them in the same semantic space, while minimizing the similarity of unpaired data:
\begingroup\makeatletter\def\f@size{8.7}\check@mathfonts\def\maketag@@@#1{\hbox{\m@th\normalfont\normalfont#1}}
\begin{align}
    L_{\mathcal{X}2\mathcal{Y}}=-\log \frac{\exp (\mathbf{x}_i^{\top} \mathbf{y}_i / \tau)}{\exp (\mathbf{x}_i^{\top} \mathbf{y}_i / \tau)+\sum_{j \neq i} \exp (\mathbf{x}_i^{\top} \mathbf{y}_j / \tau)}, \label{Eq:MolBind}
\end{align}
\endgroup
where $\tau$ is the temperature and $y_{j}$ is the in-batch negative sample~\cite{radford2021learning}. We use the symmetric loss to align $\mathcal{X}$ and $\mathcal{Y}$: $\mathcal{L}_{\mathcal{X}2\mathcal{Y}}+\mathcal{L}_{\mathcal{Y}2\mathcal{X}}$.  \textsc{MolBind} applies this loss framework to jointly bind 4 pairs of modalities, including molecular language-graph, language-conformation, graph-conformation, and conformation-protein.

\begin{table*}[t]
\centering
\scriptsize
\caption{\small Performance of cross-modal retrieval on test sets of MolBind-M4 includes the following scenarios: Graph to Language (G2L), Language to Graph (L2G), Language to Conformation (L2C), and Conformation to Language (C2L) retrieval.}
\vspace{-3mm}
\fontsize{11}{10}\selectfont
\label{tab:cross-modal-retrieval}
    \resizebox{1.0\textwidth}{!}{
 \begin{tabular}{lcccccccccccccccc}
    \toprule
    & \multicolumn{8}{c}{\textbf{Retrieval in batch}} & \multicolumn{8}{c}{\textbf{Retrieval in full test set}} \\ 
    \cmidrule(lr){2-9}  \cmidrule(lr){10-17}
    & \multicolumn{2}{c}{G2L (\%)} & \multicolumn{2}{c}{L2G (\%)}  & \multicolumn{2}{c}{C2L (\%)} & \multicolumn{2}{c}{L2C (\%)} & \multicolumn{2}{c}{G2L (\%)} & \multicolumn{2}{c}{L2G (\%)}  & \multicolumn{2}{c}{C2L (\%)} & \multicolumn{2}{c}{L2C (\%)} \\\cmidrule(lr){2-3}\cmidrule(lr){4-5} \cmidrule(lr){6-7} \cmidrule(lr){8-9} \cmidrule(lr){10-11} \cmidrule(lr){12-13} \cmidrule(lr){14-15} \cmidrule(lr){16-17} 
    Model & R@1 & R@20 & R@1 & R@20 & R@1 & R@20 & R@1 & R@20 & R@1 & R@20 & R@1 & R@20 & R@1 & R@20 & R@1 & R@20 \\\midrule

    Sci-BERT &  78.2 & 89.5 & 76.1 & 88.5  & 77.2 & 88.8  & 77.9  & 88.3 & 61.0 & 85.2 & 55.7 & 86.3 & 52.6 & 84.7 & 49.0  & 91.2 \\
    KV-PLM &  78.5 & 90.3 & 76.6 & 89.1  & 77.9 & 89.2  & 78.8  & 88.7 & 61.5 & 85.9 & 56.1 & 86.6 & 53.2 & 85.1 & 49.3  & 91.5 \\ 
    \midrule
    MoMu &    80.1 & 91.5 & 78.4 & 90.2  & 79.1 & 90.5  & 79.3  & 91.1 & 63.9 & 87.6 & 58.0 & 87.8 & 54.8 & 86.3 & 50.2  & 93.1       \\ 
    MolCA & \underline{85.3} & \underline{94.2} &\underline{82.5} & \underline{94.8}  & \underline{83.9} & \underline{94.7}  & \underline{83.5}  & \underline{94.2} & \underline{67.1} & \underline{91.8} & \underline{60.5} & \underline{91.2} & \underline{58.2} & \underline{90.7} & \underline{55.6}  & \underline{95.6}  \\
    \midrule
    \textsc{MolBind} & \textbf{88.5} & \textbf{98.6} & \textbf{86.5} & \textbf{98.7} & \textbf{86.0} & \textbf{98.6} & \textbf{85.1} & \textbf{98.5}  & \textbf{71.5}  & \textbf{93.0}  & \textbf{64.3}  & \textbf{93.6} & \textbf{62.4} & \textbf{94.0} &  \textbf{58.9}  &  \textbf{98.5} 
    
    \\
    \bottomrule
    \end{tabular}}
    \vspace{-12pt}
\end{table*}

\subsection{Dataset Collection}
In this section, we briefly describe how to construct the \texttt{MolBind-M4} dataset (full details in Appendix~\ref{app:datasets}), including 322K pairs of language-graph data, 161K pairs of language-conformation data, 161K pairs of graph-conformation data, and 72K pairs of conformation-protein data. \\
\noindent \textbf{Language-Graph Pairs:} PubChem~\cite{kim2023pubchem}, the most extensive source of chemical information, provided many molecules with diverse descriptions. We transformed these molecules into graphs and paired them with corresponding textual descriptions, yielding 322K language-graph pairs.\\
\noindent \textbf{Language-Conformation Pairs:} Using Molecule3D~\cite{xu20233d}, which contains 3.9M  ground-state molecular conformations, and the GEOM dataset~\cite{axelrod2022geom} with 37M higher-quality conformations, we matched molecular IDs (CIDs) and InChIs in them with textual descriptions from PubChem. This process resulted in 161K language-conformation pairs.\\
\noindent \textbf{Conformation-Protein Pairs:} By leveraging data from PDBBind~\cite{wang2005pdbbind} and CrossDocked~\cite{Francoeur20203DCN}, we extracted and refined conformation-protein pairs. After filtering based on RMSD thresholds and clustering for sequence identity, we consolidated our findings into 72K unique pairs, ensuring no overlap with previously identified pocket proteins.\\
\noindent \textbf{Graph-Conformation Pairs:} Converting 3D molecular structures from the language-conformation dataset into 2D graphs allowed us to generate 161K graph-conformation pairs.


\section{Experiments}
\label{sec:exp}
In this section, we evaluate the effectiveness of \textsc{MolBind} in various downstream tasks through different experiments. Firstly, \textsc{MolBind}’s capability to align molecular modalities and text is evaluated using zero-shot cross-modal retrieval and classification. In addition, we use \textsc{MolBind} to enhance the performance of downstream tasks that involve molecules and proteins. The pre-training settings can found in Appendix~\ref{app:setting}.

\subsection{Zero-shot Molecule-Language Retrieval}
\label{exp:Retrieval}
We evaluate the graph-language and conformation-language retrieval performance on the \texttt{MoBind-M4} dataset. For our baselines, we employ Sci-BERT~\cite{beltagy2019scibert}, KV-PLM~\cite{zeng2022deep}, MoMu~\cite{su2022molecular}, and MolCA~\cite{liu2023molca}, assessing their performance using  Recall@1 and Recall@20. These metrics are evaluated both within a batch of 64 samples and across the entire test set, following the methodology outlined in~\cite{liu2023molca}. As presented in Table~\ref{tab:cross-modal-retrieval}, \textsc{MolBind}'s performance surpasses that of MolCA  by a large margin. Overall, \textsc{MolBind} significantly outperforms prior work on two molecule-language retrieval tasks, validating its effectiveness in aligning the molecule and language modalities.

\begin{figure}[t]
\centering
\includegraphics[width=0.4\textwidth]{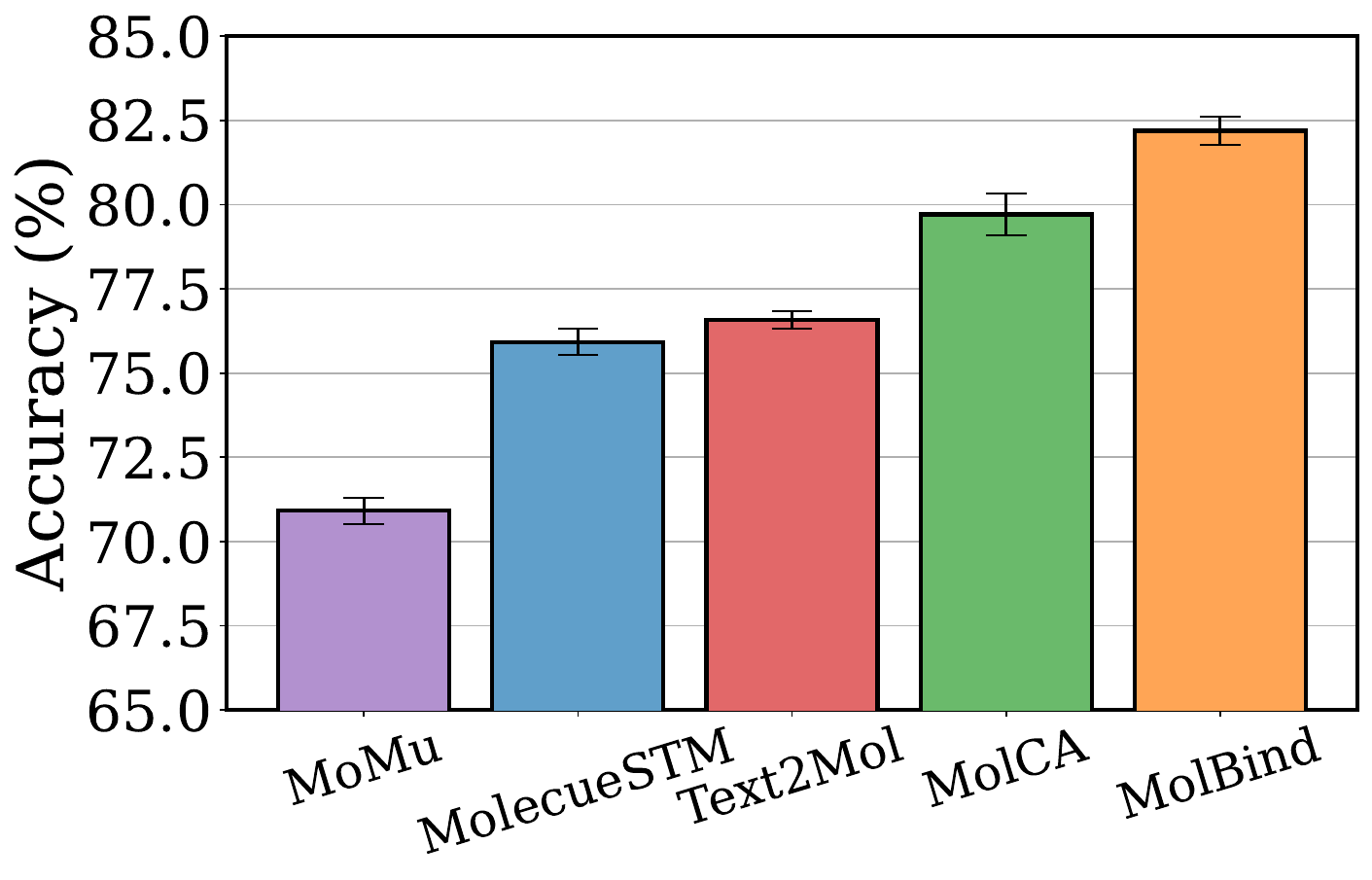}
\vskip -1em
     \caption{\small Zero-shot molecule classification with language.}
\label{fig:classification} 
\vskip -1em
\end{figure}

\subsection{Zero-shot Classification with Language}
The International Union for Pure and Applied Chemistry (IUPAC) has established a standardized naming system for chemical compounds, known as IUPAC names~\cite{favre2013nomenclature}. Notably, this system relies on identifying specific molecule structures. Therefore, correctly predicting IUPAC names indicates a model’s ability to understand molecule structures. \textsc{MolBind} has the capability for zero-shot IUPAC namultiple modalityn, thanks to its aligned representation space of molecule and language descriptions (see detailed implementation in Appendix~\ref{app:sec:class}).  For evaluation, we collect IUPAC names for molecules in valid and test sets of \texttt{MolBind-M4} using the PubChemPy library, and report average accuracy of test set.  We compare the following baselines: MoMu~\cite{su2022molecular}, MolCA~\cite{liu2023molca}, Text2Mol~\cite{edwards2021text2mol} and MoleculeSTM~\cite{liu2023multi}.

Figure~\ref{fig:classification} shows our \textsc{MolBind} consistently outperforms baselines by a large margin, showing that modeling multiple modality for molecules is an important factor in comprehending molecule structures.  These results also demonstrate the data efficiency of \textsc{MolBind}-induced zero-shot classifiers based on the aligned representation space of 2D graph and language descriptions.

\subsection{Zero-shot Molecule to Protein Retrieval}
Since \textsc{MolBind} has the ability to align the representations of molecular conformations and proteins, it can also be used for conformation-to-protein retrieval, which is known as target fishing in drug discovery. We established a benchmark using the CASF-2016 dataset. We compared \textsc{MolBind} with Vina~\cite{varadi2022alphafold}, Glide~\cite{halgren2004glide}, and DrugCLIP~\cite{gao2023drugclip}. As shown in Table~\ref{tab:retrieve_pro}, our \textsc{MolBind} significantly outperforms Vina, Glide, and DrugCLIP. More importantly, the improvement of \textsc{MolBind} over DrugCLIP, which utilizes only two modalities, validates its ability to align molecular conformations and protein modalities using multiple modalities.

\subsection{Ablation Study}
In Table~\ref{tab_ablation_vertical}, we analyze the effect of different modalities in \textsc{MolBind}'s multi-modal pretraining. When each objective function
is individually applied (1st, 2nd and 3rd row), the modality for
(language, graph) performs best on L2G, which probably
benefits from its explicit alignment. Interestingly, we observe an emergent behavior in the embedding space that aligns two pairs of modalities (language, conformation) even though we only train using the pairs (language, graph) and (graph,conformation). These results demonstrate that \textsc{MolBind} aligns the modalities and implicitly transfers the language supervision associated with conformation to other modalities such as graph. Overall, \textsc{MolBind} achieves the best performance, validating that it is an effective way to align multiple modalities simultaneously.

\begin{table}[t]
    \centering
    \setlength{\tabcolsep}{9pt}
    \fontsize{10pt}{9}\selectfont
    \caption{\small Conformation-to-protein retrieval on CASF-2016.}
    \vspace{-1em}
        \label{tab:retrieve_pro}
    \resizebox{0.49\textwidth}{!}{
    \begin{tabular}{lccccc} \toprule
                   & \multicolumn{5}{c}{Conformation to Protein (Recall)}                   \\ 
                   \cmidrule(lr){2-6} 
Model         & R@1  & R@2  & R@3  & R@4  & R@5  \\ 
\midrule
Vina         & 3.38  &  5.26  &  7.52  & 9.02  & 10.15  \\ 
Glide         & 14.98  & 22.85  &  30.34 & 35.58  & 39.33  \\ 
\midrule
DrugCLIP         & 24.07  & 42.96 &  51.11 & 59.26  & 62.59 \\ 
\midrule
\textsc{MolBind}  & \textbf{45.96}  & \textbf{48.31}  &  \textbf{60.15}  & \textbf{68.25}  & \textbf{75.92} \\ 
 \bottomrule
\end{tabular}}
 \end{table}

\section{Conclusions and Discussion}
\label{sec:concl}

\begin{table}[t]
    \centering
    \caption{\small Ablation study for using different modalities on language to graph (L2G) and language to conformation (L2C) retrieval tasks. We follow the same setting in Section~\ref{exp:Retrieval}.} \label{tab_ablation_vertical}
    \fontsize{11pt}{9}\selectfont
     \resizebox{0.49\textwidth}{!}{
    \begin{tabular}{lcc}
        \toprule[1.0pt]
        \textbf{Ablations} & \textbf{L2G} & \textbf{L2C}  \\
        \midrule
        \small{(language, graph)} & 60.4 & 0.23 \\
        \small{(language, conformation)} & 0.51 & 55.7 \\
        \small{(graph, conformation)} & 0.25 & 0.31 \\
        \small{(language, graph) \& (language, conformation)} & \underline{62.8} & \underline{58.1} \\
        \small{(language, graph) \&  (graph, conformation)} & 61.4 & 56.1 \\
        \small{(language, conformation) \&  (graph, conformation)} & 60.4 & 57.6 \\
        \midrule
       \textsc{MolBind} & \textbf{64.3} & \textbf{58.9} \\
        \bottomrule[1.0pt]
    \end{tabular}}
    \vspace{-2mm}
\end{table}

\textsc{MolBind} is a simple yet effective framework that aligns multiple modalities related to molecules in a joint embedding space. \textsc{MolBind} achieves multi-modal alignment under the constraints of insufficient biochemical data, facilitating knowledge transfer between modalities. We also release the multi-modal dataset \texttt{MolBind-M4}, which is the first dataset to combine multiple modalities, sourced exclusively from openly accessible datasets. We evaluate \textsc{MolBind} in various downstream tasks, demonstrating its effectiveness in aligning representations from multiple modalities. 



\section{Limitations and Impact Statements}
This paper presents work aimed at learning a unified joint embedding space for multiple modalities related to language and molecules. While we provide a high-quality, multi-modal training dataset, the scale of our data is comparatively smaller than that in the vision-language domain. Expanding and refining multi-modal pre-training datasets is a crucial future work for scientific research scenarios.


\bibliography{anthology,custom}

\begin{thebibliography}{43}
\expandafter\ifx\csname natexlab\endcsname\relax\def\natexlab#1{#1}\fi

\bibitem[{Axelrod et~al.(2022)Axelrod, Gomez-Bombarelli, and
  Rafael}]{axelrod2022geom}
Simon Axelrod, Gomez-Bombarelli, and Rafael. 2022.
\newblock Geom, energy-annotated molecular conformations for property
  prediction and molecular generation.
\newblock \emph{Scientific Data}, 9(1):185.

\bibitem[{Beltagy et~al.(2019)Beltagy, Lo, and Cohan}]{beltagy2019scibert}
Iz~Beltagy, Kyle Lo, and Arman Cohan. 2019.
\newblock Scibert: A pretrained language model for scientific text.
\newblock In \emph{Proceedings of the 2019 Conference on Empirical Methods in
  Natural Language Processing and the 9th International Joint Conference on
  Natural Language Processing (EMNLP-IJCNLP)}, pages 3615--3620.

\bibitem[{Bento et~al.(2020)Bento, Hersey, F{\'e}lix, Landrum, Gaulton,
  Atkinson, Bellis, De~Veij, and Leach}]{bento2020open}
A~Patr{\'\i}cia Bento, Anne Hersey, Eloy F{\'e}lix, Greg Landrum, Anna Gaulton,
  Francis Atkinson, Louisa~J Bellis, Marleen De~Veij, and Andrew~R Leach. 2020.
\newblock An open source chemical structure curation pipeline using rdkit.
\newblock \emph{Journal of Cheminformatics}, 12:1--16.

\bibitem[{Edwards et~al.(2021)Edwards, Zhai, and Ji}]{edwards2021text2mol}
Carl Edwards, ChengXiang Zhai, and Heng Ji. 2021.
\newblock Text2mol: Cross-modal molecule retrieval with natural language
  queries.
\newblock In \emph{Proceedings of the 2021 Conference on Empirical Methods in
  Natural Language Processing}, pages 595--607.

\bibitem[{Favre and Powell(2013)}]{favre2013nomenclature}
Henri~A Favre and Warren~H Powell. 2013.
\newblock \emph{Nomenclature of organic chemistry: IUPAC recommendations and
  preferred names 2013}.
\newblock Royal Society of Chemistry.

\bibitem[{Francoeur et~al.(2020)Francoeur, Masuda, Sunseri, Jia, Iovanisci,
  Snyder, and Koes}]{Francoeur20203DCN}
Paul~G. Francoeur, Tomohide Masuda, Jocelyn Sunseri, Andrew Jia, Richard~B.
  Iovanisci, Ian Snyder, and David~Ryan Koes. 2020.
\newblock \href {https://api.semanticscholar.org/CorpusID:221383064} {3d
  convolutional neural networks and a crossdocked dataset for structure-based
  drug design.}
\newblock \emph{Journal of chemical information and modeling}.

\bibitem[{Gao et~al.(2023)Gao, Qiang, Tan, Jia, Ren, Lu, Liu, Ma, and
  Lan}]{gao2023drugclip}
Bowen Gao, Bo~Qiang, Haichuan Tan, Yinjun Jia, Minsi Ren, Minsi Lu, Jingjing
  Liu, Wei-Ying Ma, and Yanyan Lan. 2023.
\newblock Drugclip: Contrastive protein-molecule representation learning for
  virtual screening.
\newblock In \emph{Thirty-seventh Conference on Neural Information Processing
  Systems}.

\bibitem[{Girdhar et~al.(2023)Girdhar, El-Nouby, Liu, Singh, Alwala, Joulin,
  and Misra}]{girdhar2023imagebind}
Rohit Girdhar, Alaaeldin El-Nouby, Zhuang Liu, Mannat Singh, Kalyan~Vasudev
  Alwala, Armand Joulin, and Ishan Misra. 2023.
\newblock Imagebind: One embedding space to bind them all.
\newblock In \emph{Proceedings of the IEEE/CVF Conference on Computer Vision
  and Pattern Recognition}, pages 15180--15190.

\bibitem[{Guzhov et~al.(2022)Guzhov, Raue, Hees, and
  Dengel}]{guzhov2022audioclip}
Andrey Guzhov, Federico Raue, J{\"o}rn Hees, and Andreas Dengel. 2022.
\newblock Audioclip: Extending clip to image, text and audio.
\newblock In \emph{ICASSP 2022-2022 IEEE International Conference on Acoustics,
  Speech and Signal Processing (ICASSP)}, pages 976--980.

\bibitem[{Hager et~al.(2023)Hager, Menten, and Rueckert}]{hager2023best}
Paul Hager, Martin~J Menten, and Daniel Rueckert. 2023.
\newblock Best of both worlds: Multimodal contrastive learning with tabular and
  imaging data.
\newblock In \emph{Proceedings of the IEEE/CVF Conference on Computer Vision
  and Pattern Recognition}, pages 23924--23935.

\bibitem[{Halgren et~al.(2004)Halgren, Murphy, Friesner, Beard, Frye, Pollard,
  and Banks}]{halgren2004glide}
Thomas~A Halgren, Robert~B Murphy, Richard~A Friesner, Hege~S Beard, Leah~L
  Frye, W~Thomas Pollard, and Jay~L Banks. 2004.
\newblock Glide: a new approach for rapid, accurate docking and scoring. 2.
  enrichment factors in database screening.
\newblock \emph{Journal of medicinal chemistry}, 47(7):1750--1759.

\bibitem[{Heller et~al.(2015)Heller, McNaught, Pletnev, Stein, and
  Tchekhovskoi}]{heller2015inchi}
Stephen~R Heller, Alan McNaught, Igor Pletnev, Stephen Stein, and Dmitrii
  Tchekhovskoi. 2015.
\newblock Inchi, the iupac international chemical identifier.
\newblock \emph{Journal of cheminformatics}, pages 1--34.

\bibitem[{Hu et~al.(2020)Hu, Liu, Gomes, Zitnik, Liang, Pande, and
  Leskovec}]{hu2020strategies}
W~Hu, B~Liu, J~Gomes, M~Zitnik, P~Liang, V~Pande, and J~Leskovec. 2020.
\newblock Strategies for pre-training graph neural networks.
\newblock In \emph{International Conference on Learning Representations
  (ICLR)}.

\bibitem[{Kim et~al.(2023)Kim, Chen, Cheng, Gindulyte, He, He, Li, Shoemaker,
  Thiessen, Yu et~al.}]{kim2023pubchem}
Sunghwan Kim, Jie Chen, Tiejun Cheng, Asta Gindulyte, Jia He, Siqian He,
  Qingliang Li, Benjamin~A Shoemaker, Paul~A Thiessen, Bo~Yu, et~al. 2023.
\newblock Pubchem 2023 update.
\newblock \emph{Nucleic acids research}, 51(D1):D1373--D1380.

\bibitem[{Kim et~al.(2019)Kim, Thiessen, Cheng, Zhang, Gindulyte, and
  Bolton}]{kim2019pug}
Sunghwan Kim, Paul~A Thiessen, Tiejun Cheng, Jian Zhang, Asta Gindulyte, and
  Evan~E Bolton. 2019.
\newblock Pug-view: programmatic access to chemical annotations integrated in
  pubchem.
\newblock \emph{Journal of cheminformatics}, 11(1):1--11.

\bibitem[{Kufareva et~al.(2012)Kufareva, Ilatovskiy, and
  Abagyan}]{kufareva2012pocketome}
Irina Kufareva, Andrey~V Ilatovskiy, and Ruben Abagyan. 2012.
\newblock Pocketome: an encyclopedia of small-molecule binding sites in 4d.
\newblock \emph{Nucleic acids research}, 40(D1):D535--D540.

\bibitem[{Kuhn et~al.(2004)Kuhn, Braslavsky, and Schmidt}]{kuhn2004chemical}
HJ~Kuhn, SE~Braslavsky, and R~Schmidt. 2004.
\newblock Chemical actinometry (iupac technical report).
\newblock \emph{Pure and Applied Chemistry}, 76(12):2105--2146.

\bibitem[{Li et~al.(2024)Li, Liu, Luo, Wang, He, Kawaguchi, Chua, and
  Tian}]{li2024towards}
Sihang Li, Zhiyuan Liu, Yanchen Luo, Xiang Wang, Xiangnan He, Kenji Kawaguchi,
  Tat-Seng Chua, and Qi~Tian. 2024.
\newblock Towards 3d molecule-text interpretation in language models.
\newblock \emph{arXiv preprint arXiv:2401.13923}.

\bibitem[{Liu et~al.(2023{\natexlab{a}})Liu, Nie, Wang, Lu, Qiao, Liu, Tang,
  Xiao, and Anandkumar}]{liu2023multi}
Shengchao Liu, Weili Nie, Chengpeng Wang, Jiarui Lu, Zhuoran Qiao, Ling Liu,
  Jian Tang, Chaowei Xiao, and Animashree Anandkumar. 2023{\natexlab{a}}.
\newblock Multi-modal molecule structure--text model for text-based retrieval
  and editing.
\newblock \emph{Nature Machine Intelligence}, 5(12):1447--1457.

\bibitem[{Liu et~al.(2023{\natexlab{b}})Liu, Li, Luo, Fei, Cao, Kawaguchi,
  Wang, and Chua}]{liu2023molca}
Zhiyuan Liu, Sihang Li, Yanchen Luo, Hao Fei, Yixin Cao, Kenji Kawaguchi, Xiang
  Wang, and Tat-Seng Chua. 2023{\natexlab{b}}.
\newblock Molca: Molecular graph-language modeling with cross-modal projector
  and uni-modal adapter.
\newblock \emph{arXiv preprint arXiv:2310.12798}.

\bibitem[{Luo et~al.(2021)Luo, Guan, Ma, and Peng}]{luo20213d}
Shitong Luo, Jiaqi Guan, Jianzhu Ma, and Jian Peng. 2021.
\newblock A 3d generative model for structure-based drug design.
\newblock \emph{Advances in Neural Information Processing Systems},
  34:6229--6239.

\bibitem[{Nakata and Shimazaki(2017)}]{nakata2017pubchemqc}
Maho Nakata and Tomomi Shimazaki. 2017.
\newblock Pubchemqc project: a large-scale first-principles electronic
  structure database for data-driven chemistry.
\newblock \emph{Journal of chemical information and modeling}, pages
  1300--1308.

\bibitem[{Peng et~al.(2022)Peng, Luo, Guan, Xie, Peng, and
  Ma}]{peng2022pocket2mol}
Xingang Peng, Shitong Luo, Jiaqi Guan, Qi~Xie, Jian Peng, and Jianzhu Ma. 2022.
\newblock Pocket2mol: Efficient molecular sampling based on 3d protein pockets.
\newblock In \emph{International Conference on Machine Learning}, pages
  17644--17655. PMLR.

\bibitem[{Radford et~al.(2021)Radford, Kim, Hallacy, Ramesh, Goh, Agarwal,
  Sastry, Askell, Mishkin, Clark et~al.}]{radford2021learning}
Alec Radford, Jong~Wook Kim, Chris Hallacy, Aditya Ramesh, Gabriel Goh,
  Sandhini Agarwal, Girish Sastry, Amanda Askell, Pamela Mishkin, Jack Clark,
  et~al. 2021.
\newblock Learning transferable visual models from natural language
  supervision.
\newblock In \emph{International conference on machine learning}, pages
  8748--8763.

\bibitem[{Rapp{\'e} et~al.(1992)Rapp{\'e}, Casewit, Colwell, Goddard~III, and
  Skiff}]{rappe1992uff}
Anthony~K Rapp{\'e}, Carla~J Casewit, KS~Colwell, William~A Goddard~III, and
  W~Mason Skiff. 1992.
\newblock Uff, a full periodic table force field for molecular mechanics and
  molecular dynamics simulations.
\newblock \emph{Journal of the American chemical society},
  114(25):10024--10035.

\bibitem[{Riniker and Landrum(2015)}]{riniker2015better}
Sereina Riniker and Gregory~A Landrum. 2015.
\newblock Better informed distance geometry: using what we know to improve
  conformation generation.
\newblock \emph{Journal of chemical information and modeling},
  55(12):2562--2574.

\bibitem[{Shuaibi et~al.(2021)Shuaibi, Kolluru, Das, Grover, Sriram, Ulissi,
  and Zitnick}]{shuaibi2021rotation}
Muhammed Shuaibi, Adeesh Kolluru, Abhishek Das, Aditya Grover, Anuroop Sriram,
  Zachary Ulissi, and C~Lawrence Zitnick. 2021.
\newblock Rotation invariant graph neural networks using spin convolutions.
\newblock \emph{arXiv preprint arXiv:2106.09575}.

\bibitem[{Sterling and Irwin(2015)}]{sterling2015zinc}
Teague Sterling and John~J Irwin. 2015.
\newblock Zinc 15--ligand discovery for everyone.
\newblock \emph{Journal of chemical information and modeling},
  55(11):2324--2337.

\bibitem[{Su et~al.(2022)Su, Du, Yang, Zhou, Li, Rao, Sun, Lu, and
  Wen}]{su2022molecular}
Bing Su, Dazhao Du, Zhao Yang, Yujie Zhou, Jiangmeng Li, Anyi Rao, Hao Sun,
  Zhiwu Lu, and Ji-Rong Wen. 2022.
\newblock A molecular multimodal foundation model associating molecule graphs
  with natural language.
\newblock \emph{arXiv preprint arXiv:2209.05481}.

\bibitem[{Varadi et~al.(2022)Varadi, Anyango, Deshpande, Nair, Natassia,
  Yordanova, Yuan, Stroe, Wood, Laydon et~al.}]{varadi2022alphafold}
Mihaly Varadi, Stephen Anyango, Mandar Deshpande, Sreenath Nair, Cindy
  Natassia, Galabina Yordanova, David Yuan, Oana Stroe, Gemma Wood, Agata
  Laydon, et~al. 2022.
\newblock Alphafold protein structure database: massively expanding the
  structural coverage of protein-sequence space with high-accuracy models.
\newblock \emph{Nucleic acids research}, 50(D1):D439--D444.

\bibitem[{Volokhova et~al.(2023)Volokhova, Koziarski,
  Hern{\'a}ndez-Garc{\'\i}a, Liu, Miret, Lemos, Thiede, Yan, Aspuru-Guzik, and
  Bengio}]{volokhova2023towards}
Alexandra Volokhova, Micha{\l} Koziarski, Alex Hern{\'a}ndez-Garc{\'\i}a,
  Cheng-Hao Liu, Santiago Miret, Pablo Lemos, Luca Thiede, Zichao Yan, Al{\'a}n
  Aspuru-Guzik, and Yoshua Bengio. 2023.
\newblock Towards equilibrium molecular conformation generation with gflownets.
\newblock \emph{arXiv preprint arXiv:2310.14782}.

\bibitem[{Wang et~al.(2005)Wang, Fang, Lu, Yang, and Wang}]{wang2005pdbbind}
Renxiao Wang, Xueliang Fang, Yipin Lu, Chao-Yie Yang, and Shaomeng Wang. 2005.
\newblock The pdbbind database: methodologies and updates.
\newblock \emph{Journal of medicinal chemistry}, 48(12):4111--4119.

\bibitem[{Xiao et~al.(2022)Xiao, Chen, Guo, Zhuang, and
  Wang}]{xiao2022decoupled}
Teng Xiao, Zhengyu Chen, Zhimeng Guo, Zeyang Zhuang, and Suhang Wang. 2022.
\newblock Decoupled self-supervised learning for graphs.
\newblock \emph{Advances in Neural Information Processing Systems},
  35:620--634.

\bibitem[{Xiao et~al.(2021)Xiao, Chen, Wang, and Wang}]{xiao2021learning}
Teng Xiao, Zhengyu Chen, Donglin Wang, and Suhang Wang. 2021.
\newblock Learning how to propagate messages in graph neural networks.
\newblock In \emph{Proceedings of the 27th ACM SIGKDD Conference on Knowledge
  Discovery \& Data Mining}, pages 1894--1903.

\bibitem[{Xu et~al.(2021{\natexlab{a}})Xu, Ghosh, Huang, Okhonko, Aghajanyan,
  Metze, Zettlemoyer, and Feichtenhofer}]{xu2021videoclip}
Hu~Xu, Gargi Ghosh, Po-Yao Huang, Dmytro Okhonko, Armen Aghajanyan, Florian
  Metze, Luke Zettlemoyer, and Christoph Feichtenhofer. 2021{\natexlab{a}}.
\newblock Videoclip: Contrastive pre-training for zero-shot video-text
  understanding.
\newblock In \emph{Proceedings of the 2021 Conference on Empirical Methods in
  Natural Language Processing}, pages 6787--6800.

\bibitem[{Xu et~al.(2021{\natexlab{b}})Xu, Luo, Zhang, Xu, Xie, Liu, Dickerson,
  Deng, Nakata, and Ji}]{xu2021molecule3d}
Zhao Xu, Youzhi Luo, Xuan Zhang, Xinyi Xu, Yaochen Xie, Meng Liu, Kaleb
  Dickerson, Cheng Deng, Maho Nakata, and Shuiwang Ji. 2021{\natexlab{b}}.
\newblock Molecule3d: A benchmark for predicting 3d geometries from molecular
  graphs.
\newblock \emph{arXiv preprint arXiv:2110.01717}.

\bibitem[{Xu et~al.(2023)Xu, Xie, Luo, Zhang, Xu, Liu, Dickerson, Deng, Nakata,
  and Ji}]{xu20233d}
Zhao Xu, Yaochen Xie, Youzhi Luo, Xuan Zhang, Xinyi Xu, Meng Liu, Kaleb
  Dickerson, Cheng Deng, Maho Nakata, and Shuiwang Ji. 2023.
\newblock 3d molecular geometry analysis with 2d graphs.
\newblock \emph{arXiv preprint arXiv:2305.13315}.

\bibitem[{You et~al.(2020)You, Chen, Sui, Chen, Wang, and Shen}]{you2020graph}
Yuning You, Tianlong Chen, Yongduo Sui, Ting Chen, Zhangyang Wang, and Yang
  Shen. 2020.
\newblock Graph contrastive learning with augmentations.
\newblock \emph{Advances in neural information processing systems},
  33:5812--5823.

\bibitem[{Zeng et~al.(2022)Zeng, Yao, Liu, and Sun}]{zeng2022deep}
Zheni Zeng, Yuan Yao, Zhiyuan Liu, and Maosong Sun. 2022.
\newblock A deep-learning system bridging molecule structure and biomedical
  text with comprehension comparable to human professionals.
\newblock \emph{Nature communications}, 13(1):862.

\bibitem[{Zhang et~al.(2022)Zhang, Guo, Zhang, Li, Miao, Cui, Qiao, Gao, and
  Li}]{zhang2022pointclip}
Renrui Zhang, Ziyu Guo, Wei Zhang, Kunchang Li, Xupeng Miao, Bin Cui, Yu~Qiao,
  Peng Gao, and Hongsheng Li. 2022.
\newblock Pointclip: Point cloud understanding by clip.
\newblock In \emph{Proceedings of the IEEE/CVF Conference on Computer Vision
  and Pattern Recognition}, pages 8552--8562.

\bibitem[{Zhou et~al.(2022)Zhou, Gao, Ding, Zheng, Xu, Wei, Zhang, and
  Ke}]{zhou2022uni}
Gengmo Zhou, Zhifeng Gao, Qiankun Ding, Hang Zheng, Hongteng Xu, Zhewei Wei,
  Linfeng Zhang, and Guolin Ke. 2022.
\newblock Uni-mol: A universal 3d molecular representation learning framework.
\newblock In \emph{The Eleventh International Conference on Learning
  Representations}.

\bibitem[{Zhu et~al.(2023)Zhu, Lin, Ning, Yan, Cui, Wang, Pang, Jiang, Zhang,
  Li et~al.}]{zhu2023languagebind}
Bin Zhu, Bin Lin, Munan Ning, Yang Yan, Jiaxi Cui, HongFa Wang, Yatian Pang,
  Wenhao Jiang, Junwu Zhang, Zongwei Li, et~al. 2023.
\newblock Languagebind: Extending video-language pretraining to n-modality by
  language-based semantic alignment.
\newblock \emph{arXiv preprint arXiv:2310.01852}.

\bibitem[{Zhu et~al.(2024)Zhu, Xiao, and Honavar}]{zhu3MDiffusion}
Huaisheng Zhu, Teng Xiao, and Vasant~G Honavar. 2024.
\newblock 3m-diffusion: Latent multi-modal diffusion for text-guided generation
  of molecular graphs.
\newblock \emph{arXiv preprint arXiv:2403.07179}.

\end{thebibliography}
\bibliographystyle{acl_natbib}

\newpage
\appendix
\section{Appendix}
\label{sec:appendix}

\subsection{The Details of Uni-Mol Encoder}
\label{sec:appencoder}
In this section, we provide additional details about the 3D encoder in Uni-Mol~\cite{zhou2022uni}. We first learn the atom embedding which is initialized using an embedding matrix based on the input atomic features of molecules $\mathbf{h}$:
\begingroup\makeatletter\def\f@size{8.7}\check@mathfonts\def\maketag@@@#1{\hbox{\m@th\normalfont\normalfont#1}}
\begin{equation}
\small
    \mathbf{m} =\mathbf{h} \mathbf{W}_{h} ,
\end{equation}
\endgroup
where $\mathbf{W}_h \in \mathbb{R}^{d \times D}$ is the learnable parameter matrix and $D$ is the hidden dimension. With the atom embedding, Uni-mol initially learns the pair-type-aware Gaussian kernel~\cite{shuaibi2021rotation} to model pairwise representation for atom pairs. This process is denoted as follows.
\begingroup\makeatletter\def\f@size{8.7}\check@mathfonts\def\maketag@@@#1{\hbox{\m@th\normalfont\normalfont#1}}
\begin{equation}
\begin{aligned}
        & \mathbf{p}_{i j}={\mathcal{G}(\mathcal{A}\left(d_{i j}, t_{i j} ; \mathbf{a}, \mathbf{b}\right), \mu^k, \sigma^k) \mid k \in[1, D]}, \\
        & \mathcal{A}(d, r ; \mathbf{a}, \mathbf{b})=a_r d+b_r,
\end{aligned}
\end{equation}
\endgroup
where $d_{i j}$ is the Euclidean distance between coordinates $\mathbf{x}_i$ and $\mathbf{x}_j$ for the atom pair $ij$, $\mathcal{G}(d, \mu, \sigma)=\frac{1}{\sigma \sqrt{2 \pi}} e^{-\frac{(d-\mu)^2}{2 \sigma^2}}$ is a Gaussian density function with parameters $\mu$ and $ \sigma$. $\mathcal{A}\left(d_{i j}, t_{i j} ; \mathbf{a}, \mathbf{b}\right)$ is the affine transformation with parameters $\mathbf{a}$ and $\mathbf{b}$. This function affines $d_{ij}$ corresponding to its pair-type $t_{ij}$. 

The pair representation is utilized to enhance the 3D spatial encoding within the model~\cite{zhou2022uni}.  It is updated via the multi-head Query-Key product results in self-attention in the layer $l$:
\begingroup\makeatletter\def\f@size{8.7}\check@mathfonts\def\maketag@@@#1{\hbox{\m@th\normalfont\normalfont#1}}
\begin{equation}
    \mathbf{q}_{i j}^{l+1}=\mathbf{q}_{i j}^l+\{\frac{\mathbf{h}_i^l \mathbf{W}_{l, h}^Q(\mathbf{h}_j^l \mathbf{W}_{l, h}^K)^T}{\sqrt{D}} \rvert\, h \in[1, H]\}, 
\end{equation}
\endgroup
where  $\mathbf{h}_i^l$ is the atom representation of the $i$-th atom at $l$-th layer and $\mathbf{q}_{ij}^l$ is the pair representation matrix of atom pair $ij$ in $l$-th layer. $H$ is the number of attention heads, $D$ is the dimension of hidden representations. $\mathbf{W}_{l, h}^Q$ and $\mathbf{W}_{l, h}^K$ are the learnable parameter matrix of the projection for Query (Key) of the $l$-th layer $h$-th head. $\mathbf{q}_{ij}^0$ is the input of the model, which is initialized with $\mathbf{p}_{ij}$. 

Following the acquisition of the pair representation, it is used by a self-attention mechanism to get the atom representations:
\begingroup\makeatletter\def\f@size{8.7}\check@mathfonts\def\maketag@@@#1{\hbox{\m@th\normalfont\normalfont#1}}
\begin{equation}
    \begin{aligned}
\mathbf{h}_i^{l+1, h} & =\operatorname{softmax}(\frac{\mathbf{h}_i^l \mathbf{W}_{l, h}^Q(\mathbf{h}_j^l \mathbf{W}_{l, h}^K)^T}{\sqrt{d}}+\mathbf{q}_{i j}^{l, h}) \mathbf{h}_j^l \mathbf{W}_{l, h}^V, \\
\mathbf{h}_i^{l+1} & =\operatorname{concat}_h(\mathbf{h}_i^{l+1, h}),
\end{aligned}
\end{equation}
\endgroup
where $\operatorname{softmax}(*)$ is the softmax function, $\operatorname{concat}_h(*)$ is the concatenation of the representation from different heads and $\mathbf{h}_i^{l+1, h}$ is the representation vector for the atom $i$ of head $h$ in the layer $l+1$. $\mathbf{W}_{l, h}^V$ is the learnable parameter matrix, which is the projection of Value of the layer $l$ and head $h$. The representation of the first layer $\mathbf{h}_i^{0}$ is initialized with the atom embedding $\mathbf{m}$. We denote the final output as $\mathbf{h}_i^{L}$ (where $L$ represents the number of layers) as $\mathbf{z}_i$, and the final representation from the 3D encoder is obtained after the mean pooling of all atom representations.

\begin{table*}[t]
\centering
\fontsize{9}{9}\selectfont
\caption{
Statistics and details of \texttt{MolBind-M4}. We randomly sample subsets for validation and testing across all modalities except for (conformation, protein), which has official validation and test sets (CASF-2016).}
\vspace{-0.5em}
\label{tab:molbind-4m}
\resizebox{0.75\textwidth}{!}{
\begin{tabular}{l c c  c c }
\toprule[1pt]
\textbf{Modality}  & \textbf{Pretrain} & \textbf{Validation} & \textbf{Test} & \textbf{Data Sources}  \\
\midrule
(language, graph)  & 319,353 & 1,500 & 1,500 & PubChem  \\
\midrule
(language, conformation) & 158,237 & 1,500 & 1,500 & Molecule3D, GEOM\\
\midrule
(conformation, protein)  & 72,355 & 100 & 285 & PDBBind, CrossDock \\
\midrule
(graph, conformation)  & 158,237 & 1,500 & 1,500 & Molecule3D  \\
\bottomrule[1pt]
\end{tabular}}
\vspace{-1em}
\end{table*}

\subsection{Details of Constructing \textsc{MolBind-M4}}
\label{app:datasets}
There are various modalities in  biology and chemistry such as language, 2D graph, 3D conformation, 3D protein (large biomolecules). In light of the varied origins of the data and potential high computational and expert annotation costs, attaining a large amount of comprehensive molecular data across all modalities is impractical. Previous studies~\cite{girdhar2023imagebind,zhu2023languagebind} have  shown that modalities can be aligned via intermediary modality even without explicitly paired data.  

Given the presence of numerous challenging-to-obtain data, especially in the field of biochemistry, such as annotations and high-quality molecular conformations, paired alignment methods are heightened. More notably, using this approach can also conveniently facilitate knowledge transfer among modalities, contributing to learning representations for modality that suffer from sparsely available data. To this end, we collected and curated four types of cross-modal molecule related pairs all from open sources to learn a single joint embedding space. In the following, we detail the construction of our \textsc{MolBind-M4}, the statistics and data source of which are provided in Table~\ref{tab:molbind-4m}.\\ 
\noindent \textbf{Language-Graph Pairs.} We used Pug View~\cite{kim2019pug} to download molecule annotations. In total, we obtained 322,353 molecules with textual descriptions. Most of these annotations begin with the common name or the International Union of Pure and Applied Chemistry (IUPAC) name~\cite{kuhn2004chemical}. To encourage the text encoder to learn generalized descriptions of molecule functionality and properties, we uniformly replaced descriptions that start with a name with 'The molecule.' We sampled high-quality validation and test sets, each consisting of 1,500 pairs with texts longer than 19 words. The remaining samples were allocated to a pre-training set, totaling 319,353 samples.\\
\noindent \textbf{Language-Conformation Pairs.} Although tools like RDKit~\cite{bento2020open} enable the generation of molecular conformations using methods such as ETKDG~\cite{riniker2015better} and empirical force fields~\cite{rappe1992uff}, these approaches often suffer from inaccuracies and poor generalizability~\cite{volokhova2023towards}. Thus, we consider ground-state conformations. 

Specifically, we initially acquired molecular conformations from two sources. The first dataset, Molecule3D~\cite{xu2021molecule3d}, includes 39M ground-state molecular conformations computed via DFT from PubChemQC~\cite{nakata2017pubchemqc}. These conformations were then matched with corresponding textual descriptions in language-graph pairs using CID as a unique identifier, resulting in 25, 961 conformation-language pairs. The second dataset, GEOM, contains approximately 37M conformations generated using a semi-empirical DFT algorithm. Due to the absence of CIDs, we matched conformation and language-graph pairs using the International Chemical Identifier (InChI)~\cite{heller2015inchi} to obtain textual descriptions and identify molecules.  Through this process, we obtained  a total of 158, 237 high-quality conformations. For data partitioning, we sampled 1,500 samples for both the training and testing sets from the most accurate ground-state conformations derived from Molecule3D. During the sampling process, we ensured that the text did not appear in the training set of language-graph pairs to mitigate potential data leakage.

\noindent \textbf{Conformation-Protein Pairs.} We initially extracted 19, 443 experimentally validated molecule-protein structure pairs from PDBBind~\cite{wang2005pdbbind}. We then expanded our dataset with the CrossDocked dataset, where structures were generated by docking ligands and pockets from the same class of binding site as delineated by Pocketome~\cite{kufareva2012pocketome}, using smina. Following previous work~\cite{luo20213d,peng2022pocket2mol}, we filtered out all poses with RMSD greater than 1Å and used MMseqs2 to cluster the data at 30\% sequence identity. We randomly selected 100K protein-ligand pairs for pre-training and another 100 pairs for the validation set. Afterward, we merged these 100K pairs with PDBBind, excluding any duplicates to form the final pre-training set, which contains 72,355 pairs. Finally, we used the widely recognized CASF2016 dataset from PDBBind as the testing dataset, which includes 285 experimentally validated structures.\\
\noindent \textbf{Graph-Conformation Pairs.} Given that 3D molecules can be converted to 2D representations using chemical tools like RDKit, we directly processed molecular structures from Conformation-Language pairs, converting them into 2D graphs. We preserved the same dataset split, incorporating 158,237 graph-conformation pairs into the pre-training set and splitting 1,500 pairs each to the training and testing sets.

\subsection{Multi-modal Pre-training Details}
\label{app:setting}
We train our model using the Adam optimizer with a learning rate of 0.001. The other hyper-parameters for encoders are set to their default values. We have a batch size of 16, and we use 4 NVIDIA A100 GPU cards for acceleration. We train our model for a maximum of 100 epochs. The temperature is set to 1 and we find that it has less effect on the performance. For more detailed training configurations, please
refer to our  code. 

\subsection{Details for Zero-shot Molecule Name Classification with Language}
\label{app:sec:class}
Given the 2D graph $G$ of  the query molecule and language $\text{“The molecule is }\{\text {IUPAC name}\}."$ with the IUPAC name $\left\{t_i\right\}_{i=1}^M$ of all $M$ names, we employ the graph encoder to extract molecule representation $\mathbf{g}$ and use the jointly learned text encoder to extract textual name representations $\left\{\textbf{t}_{i}\right\}_{i=1}^M$. We then derive classification logits $\left\{y_i\right\}_{i=1}^M$ by comparing the dot product similarity between molecule and name representations: $y_i=\mathbf{g} \cdot \textbf{t}_{i} / \tau (i=1, \cdots, M)$, which follows the formula of the InfoNCE loss in Equation~\ref{Eq:MolBind} in the main paper. Softmax is performed on these logits to derive classification probabilities over IUPAC name space.

\end{document}